\documentclass[conference]{IEEEtran}
\IEEEoverridecommandlockouts
\usepackage{gensymb}
\usepackage[numbers]{natbib}
\usepackage{cite}
\usepackage{amsmath,amssymb,amsfonts}
\usepackage{algorithmic}
\usepackage{graphicx}
\usepackage{textcomp}
\usepackage{xcolor}
\usepackage{amsmath}
\usepackage[]{algorithm2e}
\usepackage{algorithmic}
\usepackage{tabu}

\DeclareMathOperator{\taninv}{tan\,inverse}

\def\BibTeX{{\rm B\kern-.05em{\sc i\kern-.025em b}\kern-.08em
    T\kern-.1667em\lower.7ex\hbox{E}\kern-.125emX}}
\begin{document}

\title{Person Identification with Visual Summary for a Safe Access to a Smart Home }

\author{\IEEEauthorblockN{1\textsuperscript{st} Shahinur Alam}
\IEEEauthorblockA{\textit{Electrical and Computer Engineering} \\
\textit{The University of Memphis}\\
Memphis,TN, USA \\
salam@memphis.edu}
\and
\IEEEauthorblockN{2\textsuperscript{nd} Mohammed Yeasin}
\IEEEauthorblockA{\textit{Electrical and Computer Engineering} \\
\textit{The University of Memphis}\\
Memphis,TN, USA \\
myeasin@memphis.edu}
}

\maketitle
\begin{abstract}
SafeAccess is an integrated system designed to provide easier and safer access to a smart home for people with or without disabilities. The system is designed to enhance safety and promote the independence of people with disability (i.e., visually impaired). The key functionality of the system includes the detection and identification of human and generating contextual visual summary from the real-time video streams obtained from the cameras placed in strategic locations around the house. In addition, the system classifies human into groups (i.e. friends/families/caregiver versus intruders/burglars/unknown). These features allow the user to grant/deny remote access to the premises or ability to call emergency services. In this paper, we focus on designing a prototype system for the smart home and building a robust recognition engine that meets the system criteria and addresses speed, accuracy, deployment and environmental challenges under a wide variety of practical and real-life situations. To interact with the system, we implemented a dialog enabled interface to create a personalized profile using face images or video of friend/families/caregiver. To improve computational efficiency, we apply change detection to filter out frames and use Faster-RCNN to detect the human presence and extract faces using Multitask Cascaded Convolutional Networks (MTCNN). Subsequently, we apply LBP/FaceNet to identify a person and groups by matching extracted faces with the profile. The visual summary is generated using transfer learning from the VGG16 model. SafeAccess sends a visual summary to the users with an MMS containing a person’s name if any match found or as ``Unknown”, scene image, facial description, and contextual information. In addition, the daily, weekly and monthly summarized report of the past incident can be queried from the system. SafeAccess identifies friends/families/caregiver versus intruders/unknown with an average F-score 0.97 and generates a visual summary from 10 classes with an average accuracy of 98.01\%.
\end{abstract}


\section{Introduction}
In 2013, a study conducted by The Christopher \& Dana Reeve Foundation showed that 1 in 50 people are living with some form of paralysis in the USA. Visual impairment is one of the most severe types of disability among six major disabilities (physical disability, visual disability, hearing disability, mental health disabilities, intellectual disability, and learning disability). According to the World Health Organization’s, an estimated 285 million people are visually impaired around the globe. Among those, 39 million are blind, and 246 million have low vision. A project called “{\bf Cost of vision}" conducted by Prevent Blindness America (PBA) revealed that the total economic burden of eye disorders and vision loss in the United States in 2013 is 139 billion. 

The people with vision impairments and limited mobility (paralyzed partially or entirely) often face difficulties interacting with the surrounding environment. The most frustrating impact of vision loss is, it creates dependency on sighted people for navigation, finding objects, reading text/labels, detecting abnormal event and intruders, and so on. Visually impaired people mostly depend on audio and other cues to identify a person or objects. A smart house with SafeAccess will allow people to access and monitor their house from within the home as well as far away from the home. In general, people want to identify known people when inside and detect abnormal activity or intruders silently from a remote distance. One of the overheads of the existing security system is, it requires continuous monitoring of real-time videos by human observers to detect abnormal activity or intruders. Noah Sulman et al. \citep{sulman2008effective} found in a study that when the number of monitoring displays increases human performance deteriorates. They reported that a human observer missed 20\% of the event while monitoring four surveillance display. However, when they increased the number of the display window to nine missing rates rose to 60\%. 

In recent years a plethora of research were reported on navigation \citep{kao1996object}, wayfinding \citep{gude2013blind}, text reading \citep{ocr}, barcode reading, currency recognition \citep{looktel}, object recognition \citep{mapelli1997role,rublee2011orb, chincha2011finding,FindIT,KeyRinger,SonicKeyFinder,FindOne,bigham2010vizwiz,taptapsee}. Also, there are reported articles on abnormal activity detection, fall detection for elderly people, gender identification, behavioral expression detection, to assist people with visual impairments. To the best of our knowledge, there is no reported literature on an assistive solution designed to detect, identify and group human and generate contextual visual summary to provide easy and safe access and remo te monitoring of home. To fill the void, we developed SafeAccess -- an integrated system to assist people with vision impairments and limited mobility. Our main contributions are designing and implementing key functionalities of the system that include (but are not limited to): 1) Automated identification of friends/families/caregiver versus intruders/burglars/unknown 2) generating contextual visual summary from the real-time video streams 3) opening/closing entrance door remotely for the person of interest 4) smart door installation 5) utilizing computational \& network resources by change detection 6) adding voice over interaction to address accessibility issue 7) detection of face orientation and frontalization of faces. In addition to those contributions, we have addressed numerous challenges in system design, development, coding, and integration.

We assumed that people with vision impairments and limited mobility will receive help from sighted people to create Personalized Profile, to install cameras and to set up Smart Door (described in respective sections). The house will be equipped with cameras covering the critical points of a home described in the section ``Camera Installations”. A raspberry pi connected to the cameras will send video stream when a change is detected (see Change Detection) in the scene to the recognition module. The data transmission between cameras and raspberry pi will be done using intra-home Wi-Fi or wired connection based on coverage area and distance. SafeAccess can be configured to work either in a standalone or integrated mode based on computational resources. In standalone mode, face recognition unit (LBP model) resides in raspberry pi and runs with limited features. However, we assumed that a home will have an internet connection and SafeAccess will run in integrated mode. The recognition module will process the video feeds to detect the person. If a person is detected, then it will search for faces and match the faces with the personal profile. A short visual summary will be generated from facial appearance and the context. The system will broadcast the visual summary to the feedback unit. The feedback unit will send the notification via email, text, MMS and will call the listed users based on their preference. The user will be able to configure the notification frequency for an event. The system will send the scene image with the notification for further investigation and allow users to open/close the entrance door remotely or call the emergency services if they find intruders/burglars. Since MMS contains scene image, it will help to prevent fooling the system by holding a printed image of friend \& families in front of the camera. Because human eyes are good at differentiating printed image vs actual image. It might be challenging for people with vision impairments to analyze those MMS, so SafeAccess will allow them to make a phone call to the friends standing at the entrance. The system will record the video streams based on user preferences and will allow them to query summarized history. 
 
\section{Related works}
The safe access to a smart home for friends/families is a three steps cascaded process. First, we need to detect \& recognize the person from the surveillance video. Second,  generating a visual summary and then opening/closing the entrance door. In order to find the presence of a human in the surveillance video, most of the traditional system first finds objects from motion information and then classify that object as human. The classification of the detected object is performed based on any of three approaches: shape-based method, motion-based method or texture-based method. In the shape-based method, shape information such as point, blob, and boxes of moving region are extracted to classify objects using pattern recognition and computer vision technique. Local Binary Pattern \citep{ojala2002multiresolution} (LBP) is a widely used technique in the texture-based method to detect human. The other popular texture-based human detection method is Histogram of Gradient (HOG). There is a comprehensive discussion of those methods in this review \citep{paul2013human}.

The recent advancement in Machine Learning and Computer Vision, especially Convolutional Neural Network (CNN) has made the object and person detection task robust and efficient compared to the last decade. The computational complexity for training a model has reduced to a great extent with the rapid improvement in high-performance computing (HPC) and GPU. Moreover, the public image repository such as ImageNet, PASCALVOC played an important role in the success of visual recognition. Alex Krizhevsky et al. \citep{krizhevsky2012imagenet} has done ground-breaking work in object recognition using CNN from a large dataset. They trained a deep CNN with 1.2 million images from 1000 different classes with huge optimization in the parameters. The training time was reduced using Rectified Linear Unit. Krizhevsky and his colleague were more focused on building a generalized object recognition model. This approach does not provide information about the object location. Dumitru Erhan et al. reported a deep neural network based scalable object recognition model \citep{erhan2014scalable} which identifies and localize objects with a probable bounding box. Girshick et al. proposed a method R-CNN \citep{girshick2014rich}: Region with CNN features, to localize objects. It shows considerable improvement from contemporary works. They achieved tremendous success by replacing HOG features with CNN features in lower layers. Using Selective Search algorithm, they generate proposal of bounding boxes which are then passed to AlexNet \citep{krizhevsky2012imagenet} to get the labels for each bounding box. They used linear regression to localize objects more precisely. However, their approach has two major drawbacks. First, Extracting CNN features and forward pass for each region proposal is computationally very expensive. Second, three models are trained separately for CNN feature extraction, classification and generating tighter bounding box. This problem has been addressed in their later work called “Fast R-CNN” \citep{girshick2015fast}. In R-CNN, the proposed regions have a significant overlap which caused to perform the repetitive computation. The redundant computation was eliminated by sharing forward passes using Region of Interest (RoIPool). Moreover, they combined three models in one network. R-CNN and Fast-RCNN use a selective search algorithm to generate object proposal which is very slow. Shaoqing Ren et al. presented a near cost free architecture called Faster-RCNN \citep{ren2015faster} to generate region proposals. They reused CNN feature map for region proposals instead of running selective search separately. Kaiming He et al. presented a technique called mask- RCNN \citep{he2017mask} which advanced localization task to pixels level. Sermanet et al. \citep{sermanet2013overfeat} presented an integrated system called “overfeat”, where a multiscale and sliding window technique was used with ConvNet to classify, localize and detect items. They allowed five guesses to identify the correct label and position of an object since an image might be cluttered with other objects. Yi Sun et al. built ``DeepID3" \citep{sun2015deepid3}, a face recognition model from stacked convolution and inception layers. Nezami et al. developed ``Face-Cap'' \citep{nezami2018face}, an image captioning model based on facial expression.
\section{System Design and Development}
\begin{figure}[htbp]
\centerline{\includegraphics [width=\columnwidth]{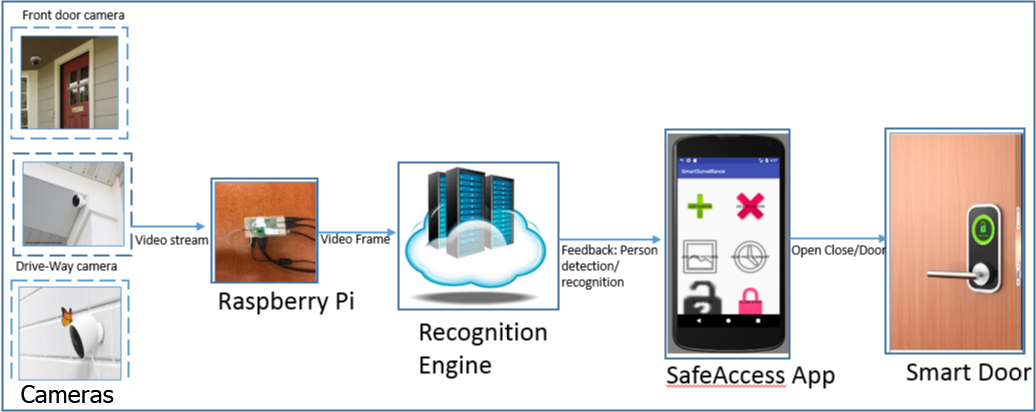}}
\caption{System Architecture}
\label{sys_arc}
\end{figure}

While designing a system, most of the assistive solutions fall short of addressing issues related to accessibility and usability. ``Design for Usability" is entirely different for people with low vision than the sighted one. Javier and colleague \citep{sanchez2012designing} introduce a concept called ``Low Vision Mobile App Portal'', which describes how to access mobile apps that are designed for visually impaired people. We considered the distribution of targeted users and their ability to receive the service and followed this \citep{UIdesign} technical guidance to design UI for creating a personal profile. SafeAccess has been compartmentalized into five modules: 1) Profile Creation module 2) Camera installation Module, 3) Smart Door setup module 4) Recognition Engine 5) Feedback module. We designed and developed each component individually, and then we integrated all pieces together by applying the learnings from ``System Thinking". ``System Thinking" is a design principle that emphasizes considering individual components of a system as a whole rather than an isolated part.  The architecture of SafeAccess is shown in figure \ref{sys_arc} and each module has been described in their respective sections.
A Raspberry Pi connected to the cameras placed at different points around the house is responsible for capturing frames and transmitting it to the recognition module. Recognition engine detects and recognizes the person and generates a visual summary. Feedback module sends the visual summary to users, and the users respond to the door using the SafeAccess app.

\subsection{Profile Creation module}
In order to identify a person, the recognition model needs to be trained with face images of friends/families/caregiver from a personalized profile. The personalized profile will contain demographic information (Name, Email, Contact, Address) and face images with different expression (Joy, Sad, Surprise, Fear, Contempt, Disgust). SafeAccess app will enable the user to collect images from camera preview/photo gallery/video clip. We have included four utility features to create a personal profile: 1) Add Person: This option is to include a new person with demographic information and images from different view 2) Add views: using this option new picture of a person can be added to the profile. Users are suggested to add multiple views of a face in the training sample so that the system can recognize a person with various poses and view angles in natural settings. 3) Delete person: this option can be used to delete all information about a person from the profile 4) Read out Summary: This option will be used to know the history of abnormal activity or appearance of unknown persons especially when a user misses any feedback. 

\subsection{Camera Installation Module}
The key factors that need to be considered before purchasing and installing cameras are 1) the durability of the camera considering the weather and temperature of the surroundings throughout the year. 2) image quality and data transmission rate:-the images with high resolution are better for image analysis and provide robust recognition outcomes, but transmission latency is high. We need to draw a trade-off between camera resolution and data transmission latency. 3) the field of view angle (FOV) 4) wireless versus wired:- wireless cameras are easy to install but when the distance between the camera and Wi-Fi hub increases the signal strength degrades. So, we need to choose a wireless versus wired camera based on the coverage area. 5) How many cameras are required:- the number of cameras required to protect a home depends on the size of the house, coverage area, and indoor versus outdoor layout. 6) Where to place: identifying critical places to install cameras is essential because monitoring the entire area is expensive. In 2005, the law enforcement agency reported more than 2 million burglary offenses in the USA \citep{burglerentry}. The survey revealed that 81\% burglar entered home through the first floor (34\% burglar chose the front door, 23\% first-floor window, 22\% back door, and 2\% storage area). Remaining 9\% burglar selected garage, 4\% basement, 4\% unlocked entrance, and 2\% anywhere on the second floor to enter inside in the house. The statistics show the first floor is more vulnerable to the burglars compare to any other point of a house. Considering theft and burglary statistics, camera can be installed at the front door, back door, off-Street windows, driveways, porches, and stairways. 4) what is the optimal view angle: How wide or narrow monitoring view we want depends on the field of view angle (FOV). If the FOV of a camera is large, it can capture a wide view but the objects in those views will be very small. If the FOV is small, it can capture a small area, but objects in those views will be large. 7) and lastly, How much it will cost.

We assumed that a camera will be installed at the entrance door at a height between 6-7 ft in addition to other locations. It will help to acquire frontalized faces of the entering person. There is a chance that people might tamper the camera if we put in such a low height. However, there are some hidden cameras available in the market those can be used just for entrance doors. The images captured from a camera located at the top- corner of a building usually has a tilted view and most of the time it is difficult to align those faces/images. The state-of-art face alignment algorithm can frontalize faces with a limited angle.
\subsection{Smart Door setup module}
\begin{figure}[htbp]
\centerline{\includegraphics [width=\columnwidth]{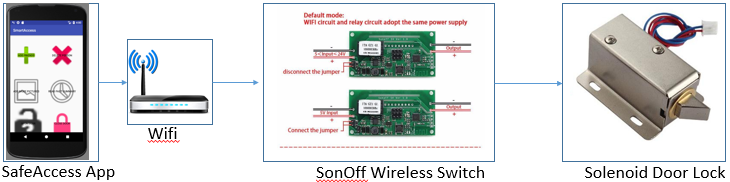}}
\caption{Smart door architecture}
\label{smartdoor}
\end{figure}

One of the vital components of a smart home is a safe and secure door with a smart lock. A smart lock allows the user to enter a house without requiring physical keys. Bluetooth and Wifi-enabled locks such as August Smart Lock \citep{augustlock}, Schlage Sense \citep{schlage}, ZigBee Lock \citep{park2009smart} etc. are primarily used to control the door. Bluetooth-enabled locks allow users to control a smart door from a short distance. On the other hand, Wifi-enabled locks are used to open/close the door from a remote distance. In our experiment, we have used Sonoff SV Wifi enabled-switch to control Solenoid door lock. The underlying architecture of the smart door is shown in Figure \ref{smartdoor}. Using SafeAccess app when users send a command to open the door then Sonoff switch turn on the lock and door will be open. The door will be closed automatically after a certain time elapsed or based on user command. This time interval can be set according to users preferences. Solenoid lock is configured to work in inverted mode. Therefore, when there is no power in the lock, the door will be closed that's how we can save power consumption. The power requirement for Sonoff varied from 5V to 250V. It supports a various generation of Wifi and the supported range of frequency is 2.4GHz to 433MHz. To ensure the highest level of security Sonoff requires custom firmware to set user-defined wifi SSID (Service Set Identifier) and password. Two widely used firmware are ESP Easy \citep{espy} and Tasmota \citep{tasmota}. The required auxiliary tools for installing custom firmware are a 0.1-inch straight pin header, USB 2.0 to TTL, solder, Arduino IDE and firmware code. To make Sonoff accessible from any distance we need to configure and add it to the home internet router.
\subsection{Recognition Engine}
\begin{figure}[htbp]
\centerline{\includegraphics [width=\columnwidth]{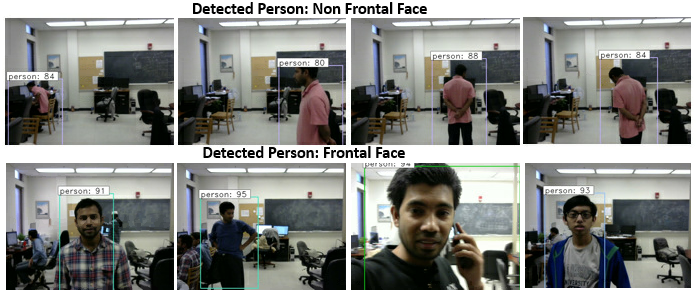}}
\caption{Person with frontal (down) \& non-frontal faces (up)}
\label{frontalnonfrontal}
\end{figure}

The Recognition module runs on a remote server and consists of two components: 1) Person Detector 2) Person Recognizer. The reason for including a person detection model (Faster R-CNN) is to notify users about human presence even if there is no face found especially when the front view of a person is not visible to the cameras. Faster R-CNN introduced Region Proposal Network (RPN) with a massive improvement in computational complexity to find object proposals by sharing convolutional layers. They reused feature map which already calculated in the forward pass of R-CNN. RPN is a fully connected convolutional layer which runs on top of CNN convolutional layers. RPN slides a small window over the CNN feature map and outputs k potential bounding boxes and associated scores indicating how likely that box will contain an object. The class labels for those proposed bounding boxes are obtained from Fast R-CNN. An image might have multiple bounding boxes with various objects and persons. In order to recognize a person, first, we need to make sure there is a face in that copped bounding box. Figure \ref{frontalnonfrontal} shows some bounding boxes with the frontal/non-frontal faces.
We used Multitask Cascaded Convolutional Networks (MTCNN) \citep{zhang2016joint} face detector to find faces in those bounding boxes. Once the face is found then it is sent to the person recognizer model to obtain a person name. We analyzed the recognition accuracy using four methods: Eigenface \citep{turk1991face}, FisherFace \citep{turk1991face}, Local Binary Pattern (LBP) \citep{ahonen2004face}, and FaceNet \citep{schroff2015facenet}. EigenFace is a Principal Component Analysis based model which projects the higher dimensional data into lower dimensional space. The drawbacks of these methods are it finds the directions with the highest variance and very error prone to noise. Some discriminative information may be lost because it does not consider classes. Fisherface is a Linear Discriminative Analysis (LDA) based method which performs class-specific dimension reductions. It maximizes between class ratios and minimizes inner class ratio. Fisherface works fine with the constrained environment and sufficient training sample. However, real-world scenarios are not perfect and have a limited option to control geometric and photometric information. LBP works reasonably well even with a minimal number of training samples. LBP extracts features such as textures \& shapes from the small local region of an image and has low dimension inherently. It calculates a binary code for a central pixel based on the intensity of surrounding pixels by applying a threshold. Then it calculates a histogram of those patterns and uses nearest neighbor classifiers to find a class label. The deep neural network has made face recognition task more robust. Parkhi et al. presented ``Deep face recognition" \citep{parkhi2015deep}; a Convolutional neural network based model to recognize faces. Florian et al. presented ``FaceNet" \citep{schroff2015facenet}, a CNN based unified embedding for face recognition. The major drawback of the deep neural network based model is that it takes a long time to train the model for low form factors device like raspberry pi. So, if the personal profile gets changed by the addition or deletion of any person or views images then retraining those model is very burdensome. We have used LBP as a primary face recognition model for Raspberry Pi in standalone mode and FaceNet for devices with high computational resources in integrated mode. 
\subsection{Feedback Module}
The primary task of the feedback module is sending a notification to the users with a visual summary. Although smartphones are easily accessible nowadays, lots of people with disability do not know how to utilize accessibility features such as TalkBack, Siri, etc. properly. Hence, designing an effective feedback system for people with disabilities is very challenging. Considering the technical adaptability of the users, we have included three types of feedback modes MMS, email, and phone call. Those feedback mode can be set based on user preference. We are using SMTP (Simple Mail Transfer Protocol) server to send those SMS \& MMS to the users via their phone operator. To make a phone call, we are using Twilio \citep{Twilio} 3rd party service which will cost \$0.013/minute. The additional task of the feedback module is storing the history of the activities in persistent storage. We have used a MySQL database to store personal profile and event history.

\section{Visual Summary Generation}
\begin{figure}[htbp]
\centerline{\includegraphics [width=\columnwidth]{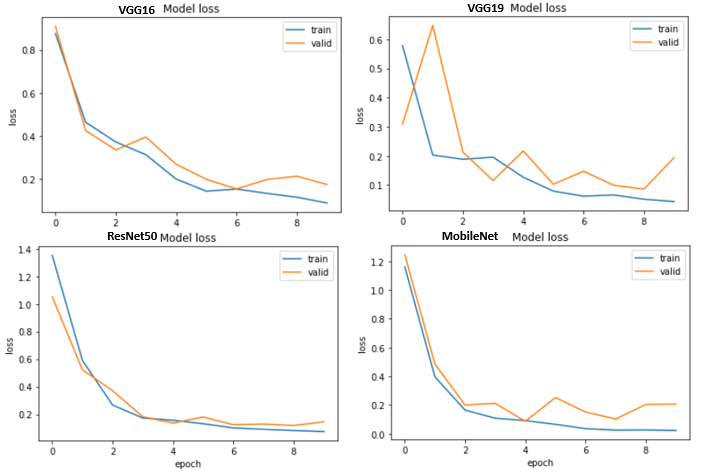}}
\caption{Training/Validation loss after transfer learning: VGG16,  VGG19,  ResNet50,  MobileNet (top left, top right,bottom left,bottom right)}
\label{trainloss}
\end{figure}

SafeAccess will generate context oriented visual summary from the video streams based on the detection and recognition outcome. The summary will include information about identified person's name if any match found, the location of that person around the house, whether he/she has a gun at hand or talking over the phone, and facial summary such as ( beard, mustache, eyeglasses, and bald head). Here is two sample visual summary: 1) ``John at entrance talking over the phone"; 2) ``An unknown person with beard/mustache/eyeglass/bald head/gun at the back door". The person recognition process is described in section D (Recognition Engine). The location of the detected person is obtained based on the source camera of the video frames. To generate the other visual summary we have used CNN with transfer learning. Transfer learning is a useful technique when we have small dataset since the deep neural network requires a large number of sample to train it. We have evaluated transfer learning with four base model (VGG16  \citep{simonyan2014very}, VGG19, ResNet50 \citep{he2016deep}, MobileNet \citep{howard2017mobilenets}) pre-trained on ImageNet dataset. The lower layers of CNN learns detailed level feature such as lines, blobs, edges, etc. whereas layers on upper level learn higher level abstracts such as information about shapes and small components of an object. We have chopped off the output layer from the base model and added eight dense layers. The base model will work as a feature extractor, and the upper-level dense layer will learn more complex class specific details. We have fine-tuned the network for five epochs with a very low learning rate (0.001) using stochastic gradient descent (SGD) optimizer. Figure \ref{trainloss} shows the loss curve for different models. From the experiments, we have found that if we increase the number of epochs more than five, then it over-fits VGG16, VGG19, and MobileNet hence validation loss starts increasing.
 
\begin{figure}[htbp]
\centerline{\includegraphics [width=\columnwidth]{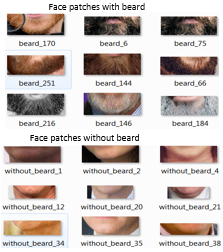}}
\caption{Face patches  }
\label{facepart}
\end{figure}
\subsection{Data collection and Pre-processing} 
We have included ten classes (cellphone, gun, eyeglass, eyes without glass, beard, non-beard, mustache, non-mustache, bald head,non-bald head) to generate context oriented visual summary and collected 5324 image samples from ImageNet \citep{deng2009imagenet}, RGB-D\citep{lai2011large} and web. Then the collected images are sorted out based on the availability of human faces with/without a beard, eyeglasses, mustache and bald head. Afterward, those face parts mentioned above have been cropped from each of the collected face images and placed similar parts in one class. For example, all face parts/patches with a beard are kept in one group and face patches without beard are in a separate group. Two sample classes/groups with few face patches are shown in figure \ref{facepart}. We wrote a program to crop/segment different parts of a face by finding facial landmarks using the technique presented by Kazemi at el. \citep{kazemi2014one}. A sample face image with landmarks and associated numbers are shown in figure \ref{frontalization}. The detailed logic of cropping face parts/patches have been presented in algorithm \ref{alg:crop_face}. The algorithm takes an image as an input and outputs four patches of a face such as an eye(ep), beard(bp), mustache(mp), head(hp). The variable ``l" contains facial landmark points and (x,y,w,h) are the bounding box of the detected face. The other methods are self-explanatory. After isolating those face parts into different groups, we manually examined every single patch and filtered out patches that are too small in size (less than 20x20) and do not have enough distinguishing patterns. To make the model affine(rotation, translation, shear, scale) invariant within a certain range we have applied data augmentations on the processed samples so that it can recognize face parts robustly with various orientation and head poses.

\begin{algorithm}[h]
\begin{algorithmic}
\caption{Crop face Patches}
\label{alg:crop_face} 
\REQUIRE \textbf{Input: } $face\_image$
\STATE \textbf{Output: } $face\_patches$
\STATE $l  \leftarrow find\_face\_landmarks(face\_image)$
\STATE $(x, y, w, h) \leftarrow find\_face\_bound\_rect(face\_image)$ 
\STATE $(x1,y1) \leftarrow l[0]$
\STATE $(x2,y2) \leftarrow l[16]$
\STATE $(x3,y3) \leftarrow l[29]$
\STATE $(x4,y4)\leftarrow l[19]$
\STATE $(x5,y5)\leftarrow l[24]$
\STATE $(x6,y6)\leftarrow l[4]$
\STATE $(x7,y7)\leftarrow l[12]$
\STATE $(x8,y8)\leftarrow l[8]$
\STATE $(x9,y9)\leftarrow l[30]$
\STATE $ep \leftarrow face\_image[y4:y3,x1:x2]$
\IF {$y-180  \leq  0 $}
\STATE $y \leftarrow 0$
\ELSE
\STATE $y \leftarrow y-180$
\ENDIF
\STATE $hp \leftarrow face\_image[y:y5,x:x2]$
\STATE $bp \leftarrow face\_image[y6:y8,x6:x7]$ 
\STATE $mp \leftarrow face\_image[y9:y6,x6:x7]$
\STATE $face\_patches \leftarrow (ep,hp,bp,mp)$    
\STATE  $return face\_patches$ 
\end{algorithmic}
\end{algorithm}

\section{System’s Novelty Features and Challenges}
The robust detection and recognition of person/intruder from surveillance video are more challenging than detecting arbitrary objects for several factors. First of all, the lighting conditions of outdoor environment changes with the time of a day and weather which sometimes makes the surveillance videos difficult to read and understand. The other factors are background clutter, facial appearance (pose, hairstyle, makeup, mustache, beard, aging, and expression), position, resolution, and field of view of a camera, etc. To overcome some of those challenges, we have included the following novelty features in SafeAccess. The rationale for incorporating those features has been described in their respective sections.
\subsection{Face frontalization}
\begin{figure}[htbp]
\centerline{\includegraphics [width=\columnwidth]{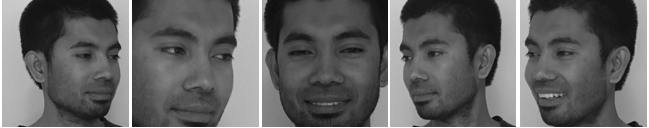}}
\caption{Faces of a person from multiple views}
\label{multview}
\end{figure}

\begin{figure}[htbp]
\centerline{\includegraphics [width=\columnwidth]{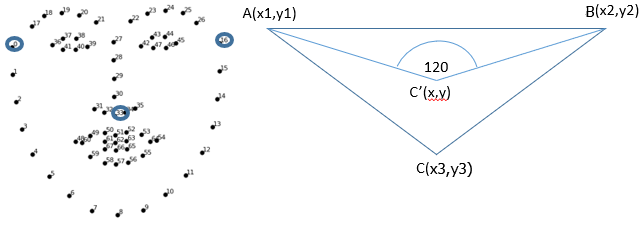}}
\caption{Facial landmarks in left image, }
\label{frontalization}
\end{figure}

\begin{figure}[htbp]
\centerline{\includegraphics [width=\columnwidth]{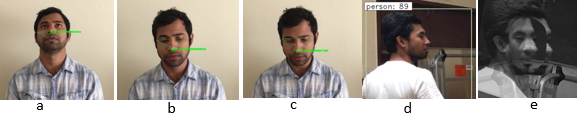}}
\caption{Sample faces with calculated tilt. a.162.11$\degree$ b.115.23$\degree$ c.93.67$\degree$}
\label{frontsample}
\end{figure}

Face frontalization is a challenging problem for any face recognition based systems because face might have various poses and orientations. Non-frontal and out of plane face is hard to recognize for any computer vision based engine. To deal with this problem, we have included faces from different views (a sample is shown in figure \ref{multview}) in the training data. 
However, it is very burdensome to include faces in the profile covering all possible orientations and pose. Face frontalization or alignment technique will increase recognition performance. The idea of face frontalization is making the eyes, lips, and nose are centered as much as possible. In order to achieve that, first, we identified faces that need to be frontalized because if we apply frontalization on faces that are already aligned it may change facial properties. We have presented a novel and straightforward approach to detect non-frontal and out of plane faces by finding 68 facial landmark point such as chin, eyes, and eyebrows (shown in figure \ref{frontalization} left) using the technique presented by Vahid Kazemi \citep{kazemi2014one}. The face orientation is detected by measuring two parameters $\alpha,\beta$; rotation about the x-axis and z-axis respectively. The parameter $\beta$ is calculated from the slope of the connecting line between the centroid of the left \& right eyes. The parameter $\alpha$ is calculated by forming a triangle with points A=0, C=33, B=16 (shown Figure \ref{frontalization} (left), points are marked with a circle). Experimentally we have found that a face with no tilt forms a triangle where angle ACB is approximately 120\degree. If the angle is above 120$\degree$ the face is tilted up and if it is below 120$\degree$ then face is tilted down. Some sample faces with different pose \& tilt angle are shown in Figure \ref{frontsample} (a,b,c).
The angle ABC (between line AC \& BC) has been calculated using equation \ref{eq:1}
\begin{equation}\label{eq:1}
\theta=\taninv\frac{m1-m2}{1+m1*m2}
\end{equation}
where m1 \& m2 are the slopes of line AC \& BC respectively. Now, based on $\alpha$ \& $\beta$, the system decides whether a face requires frontalization or not and apply ``Effective face frontalization in unconstrained images" techniques presented by Hassner et al. \citep{hassner2015effective}  for faces with extreme postures only. Figure \ref{frontsample} (d \& e) shows a sample face with extreme posture and  corresponding frontalization outcome. The average recognition accuracy after applying face frontalization on faces with large tilt and skewd orientation is 23\% while systems fails completely to identify those faces without frontalization. 

\subsection{Guided creation of Personal Profile} 
\begin{figure}[htbp]
\centerline{\includegraphics [width=\columnwidth]{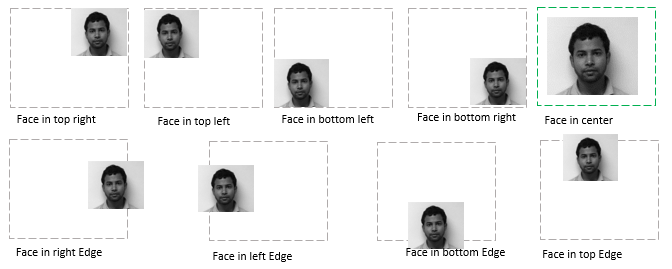}}
\caption{Guidance on profile creation: face positions in camera preview}
\label{guide_prof}
\end{figure}

In order to facilitate personal profile creation (Section A above), we have included this feature. When the users select "Add Person" or "Add views" from the SafeAccess App, the system open camera in preview mode. A face recognizer has been incorporated to make the image acquisition convenient. It will make sure selected view has face in right position by providing feedback such as "Face in top right", "Face in top left", "Face in bottom right", "Face in bottom left”, “Face in left Edge", "Face in right Edge", "Face in top Edge", "Face in bottom Edge” and "Face in center" (sample shown in \ref{guide_prof}). If the face is found near the edges of the camera window, it is better to change the position of camera or person to bring the face in the center of the window. Moreover, if the person is far away from the camera and the size of the detected face is very small then the system guides them to come closer. We calculated those aforementioned positions of the face based on the four corners of the bounding box of detected faces. The underlying logic is shown in algorithm \ref{alg:guide_prof_alg} where x, y (top left corner), ``width'', and ``height'' are obtained from face-bounding box.

\begin{algorithm}[h]
\begin{algorithmic}
\caption{Guidance to capture face images}
\label{alg:guide_prof_alg} 
\REQUIRE \textbf{Input: }  $frame$
\STATE \textbf{Output: } $face\_position$
\STATE ($x, $y, $width, height) \leftarrow get\_face\_bounding\_box(frame)$
\ENSURE $x > 0, y > 0, width > 0, height > 0$
\STATE $x1,$y1,$x2,$y2,$x3,$y3,$x4,$y4
\STATE $w \leftarrow $frame.width , $h  \leftarrow $frame.height
\STATE $x1 \leftarrow $x - $width/2 $
\STATE $y1 \leftarrow $y - $height/2 $
\STATE $x2 \leftarrow $x1 + 3 * $width/2 $
\STATE $y2 \leftarrow $y - $height/2 $
\STATE $x3 \leftarrow $x - $width/2 $
\STATE $y3 \leftarrow $y + 3 * $height/2 $
\STATE $x4 \leftarrow $x + 3 * $width/2 $
\STATE $y4 \leftarrow $y + 3 * $height/2 $

\IF {$width * height  \leq  1024 $}
\STATE $face\_position \leftarrow \textsl{"Face is small. come closer"}$

\ELSIF {$x1 \leq  0 \land  $y1 $\leq  0 $}
\STATE $face\_position \leftarrow \textsl{"Face in top left"}$
\ELSIF {$x2 \geq  w \land  $y2 $\leq  0 $}
\STATE $face\_position \leftarrow \textsl{"Face in top right"}$
\ELSIF {$x3 \leq  0 \land  $y3 $\geq  h $}
\STATE $face\_position \leftarrow \textsl{"Face in bottom left"}$
\ELSIF {$x4 \geq  w \land  $y4 $\geq  h $}
\STATE $face\_position \leftarrow \textsl{"Faci in bottom right"}$
\ELSIF {$x1 \leq  0 $}
\STATE $face\_position \leftarrow \textsl{"Face in left edge"}$
\ELSIF {$y1 \leq  0 $}
\STATE $face\_position \leftarrow \textsl{"Face in top edge"}$
\ELSIF {$x2 \geq  w $}
\STATE $face\_position \leftarrow \textsl{"Face in right edge"}$
\ELSIF {$y4 \geq  h $}
\STATE $face\_position \leftarrow \textsl{"Face in bottom edge"}$
\ELSE 
\STATE $face\_position \leftarrow \textsl{"Face in center"}$
\ENDIF

\end{algorithmic}
\end{algorithm}
 
\subsection{Change Detection}
\begin{figure}[htbp]
\centerline{\includegraphics [width=\columnwidth]{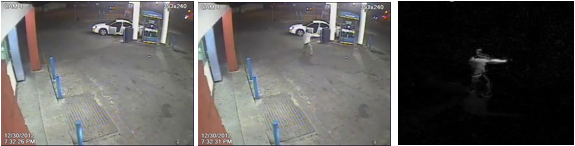}}
\caption{Change detection from two frames}
\label{change_det}
\end{figure}

\begin{figure}[htbp]
\centerline{\includegraphics [width=\columnwidth]{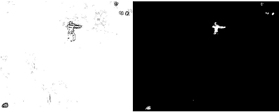}}
\caption{Change detection with adaptive (left) and binary (right) threshold}
\label{threshold}
\end{figure}

Although nowadays computational resources are easily accessible, the fast processing unit like GPU or HPC is still expensive.  Considering the cost-effectiveness and network utilization, we only transmit, process and store a video frame that has an activity or a change has been detected. The captured camera stream will not have any activity unless an event occurs such as a person enters into the monitoring area or any natural events like rain, storm changes the scene. In order to find a frame with activity, we perform change detection by subtracting consecutive frames. In figure \ref{change_det}, we have shown the outcome of change detection (right) from two frames (left, middle). In the middle frame an event, shooting is happening. 
The outcome of the change detection can be affected by the changes in contrast, brightness, and unwanted artifacts. Using gamma correction we neutralized the effect of the brightness on change detection. The other unwanted artifact were suppressed by applying two levels of thresholding; 1) Pixel level thresholding: each pixel on the subtracted frame is compared with a threshold if the pixel is above the threshold then assigning a value 255. We examined the robustness of change detection with two types of pixel level threshold; a) binary or global user-defined threshold b) adaptive threshold with a Gaussian window. The only difference between adaptive and binary thresholding is that one learns the threshold from neighborhood pixels, and other has a global predefined threshold. We found from the experiment is that binary thresholding provides robust change detection compare to adaptive. Adaptive thresholding is more sensitive to noise, contrast \& brightness. Figure \ref{threshold} shows a comparison of two approaches. 
2) Global thresholding: we applied a global threshold on summed pixels value for the entire subtracted frame on top of pixel level threshold. We assumed for robust identification face size would be at least 32X32. So total changes in pixels values would be at least 32*32*255=261120 in two successive frames when a person appears in the scene. If we set this threshold to find activities/changes the model becomes very conservative hence generates less false positive (high precision 0.991) and high false negative (low recall 0.94). Provided the context we are dealing with; a high recall is more important to us compare to high precision since we don't want to miss any activity. We analyzed the reason for generating high false negative and found that there some frames where the spread of the changes exceeds the region 32X32, and pixel difference is zero inside that region due to the first level threshold. However, when we reduced the threshold to less than 100000 model becomes very liberal and fails to filter out frames that do not have significant activities. So we considered 100000 as the second level threshold

\subsection{Remote control of Smart Door} SafeAccess will allow the users to see and verify the recognized person and remotely open the door. The user will be able to open the door from anywhere for their family \& friends as long as they have an internet connection. This option will eliminate the necessity of carrying physical keys. 

\section{Model Training} The person detection model, Faster-RCNN has been trained with PASCALVOC 2012 dataset which contains 11,530 images. The size of the input image is 600x600. NVIDIA GTX 1080 GPU with 2560 CUDA core has been used to train the model for 300 iterations with batch size 600. It took 74 hours to train the ResNet50. The person recognizer model was trained with 180 images captured from 16 subjects and it took one minute 55 seconds to train. The visual summary generator consumed 15 minutes for transfer learning and fine-tuning VGG16.  

\begin{table}[htbp]
\centering
\caption{Average F-measure of person identification}
\label{tab:recog}
\begin{tabular}{|c|c|} \hline
Model&Average F-measure of person identification \\ \hline
EigenFace & 0.94 \\ \hline
FisherFace & 0.95 \\ \hline
LBP & 0.96 \\ \hline
FaceNet & 0.97 \\ \hline

\hline\end{tabular}
\end{table}
\section{ Experiment and results}SafeAccess has been evaluated with real-time video streams collected using Logitech C270 HD webcam placed in-front of the door. We recorded each video session so that we can evaluate different recognition model with identical settings. Those sessions were captured at a different time point of a day to check how the system performs at different lighting conditions. The average F-measures of identifying friends/families using different models are shown in table \ref{tab:recog}. Among the four models, we have selected LBP and FaceNet because of their robustness. We found some scenarios where SafeAccess failed to identify persons when faces were out of the plane with a large tilt. 
\begin{table}[htbp]
\centering
\caption{Average accuracy of generating visual summary}
\label{tab:vis_sum}
\begin{tabular}{|p{1.5cm}|p{1.5cm}|p{1.5cm}| p{1.5cm}} \hline
Base Model & patch size(20x20 to 32x32) & patch size(32x32 to 64x64) & patch size 64x64 to up \\ \hline
VGG16 & 85\% & 94\% &98.01\% \\ \hline
VGG19 & 84\% &92\%  & 97.07\% \\ \hline
ResNet50 & 82\% &90\%  & 96.00\% \\ \hline
MobileNet &  75\% &88\%  & 96.10\% \\ \hline
\hline\end{tabular}
\end{table}
The classification accuracy of generating visual summary using transfer learning from different models is shown in table \ref{tab:vis_sum}. We can see from the table is that when the size of the face patch/part is very small classification accuracy is very poor. We analyzed some of those tiny patches, and it was hard to understand even with human eyes. However, the model provided a very robust classification outcome with an average accuracy of 98.01\% for the face patches of size greater than or equal 64x64. Three cascaded models take less than a second to detect, recognize and generate the visual summary.

\section{Conclusion}  In this paper, we have presented SafeAccess, an interactive assistive solution to monitor who is entering \& leaving home and identify friends/families/caregiver vs intruders/unknown. SafeAccess provides safe and easier access to a smart home and helps people with disability to live independently and with dignity. Here I will quote comments from Apple senior manager, Sarah Herrlinger: “For some people, doing something like turning on your lights or opening a blind or changing your thermostat might be seen as a convenience, but for others, that represents empowerment, and independence, and dignity” \citep{appmanager}. In the alpha version, we have designed a prototype for a smart house and built a robust model to generate a visual summary using transfer learning. In the beta version, we will 1) incorporate more sophisticated technique to frontalize faces 2) develop an algorithm to remove/neutralize the effect of rain in the video streams. 3) add servo motor to adjust the camera mount to provide the best possible coverage when any activity found.

\bibliographystyle{IEEEtran}
\bibliography{conference_041818}

\end{document}